%% file: Main.tex
\documentclass[twocolumn, switch]{article}
\usepackage{preprint}
\usepackage{algorithm}
\usepackage{algorithmic}
\usepackage{amsmath, amsthm, amssymb, amsfonts}

\usepackage[numbers,square]{natbib}
\usepackage[utf8]{inputenc}
\usepackage[T1]{fontenc}
\usepackage{xcolor}
\usepackage[colorlinks = true,
            linkcolor = purple,
            urlcolor  = blue,
            citecolor = cyan,
            anchorcolor = black]{hyperref}
\usepackage{booktabs}
\usepackage{nicefrac}
\usepackage{microtype}
\usepackage{lineno}
\usepackage{float}
\usepackage{lipsum}
\usepackage{newfloat}
\DeclareFloatingEnvironment[name={Supplementary Figure}]{suppfigure}
\usepackage{sidecap}
\sidecaptionvpos{figure}{c}

\usepackage{titlesec}
\titlespacing\section{0pt}{12pt plus 3pt minus 3pt}{1pt plus 1pt minus 1pt}
\titlespacing\subsection{0pt}{10pt plus 3pt minus 3pt}{1pt plus 1pt minus 1pt}
\titlespacing\subsubsection{0pt}{8pt plus 3pt minus 3pt}{1pt plus 1pt minus 1pt}

\usepackage{eso-pic}
\usepackage{tikz}
\usepackage{xcolor}
\PassOptionsToPackage{colorlinks=true,linkcolor=gray,urlcolor=gray}{hyperref}

\usepackage{titling}
\usepackage{footmisc}
\usepackage{multirow}
\usepackage{amsmath}

\newcommand{\Author}[2]{\textbf{#1}\textsuperscript{#2}}

\title{SlideMix: Enhancing Whole Slide Image Analysis via Multimodal Shuffling}

\author{
  \Author{Chad Wong}{1} \and
  \Author{Sicheng Chen}{1} \and
  \Author{Tianyi Zhang}{2} \and
  \Author{Enhui Chai}{3} \and
  \Author{Yueming Jin}{2} \and
  \Author{Zeyu Liu}{3} \and
  \Author{Fei Xia}{1}
}

\date{
\textsuperscript{1}Nhu Department of Electrical Engineering and Computer Science, University of California, Irvine \\
\textsuperscript{2}Department of Electrical \& Computer Engineering, National University of Singapore \\
\textsuperscript{3}PuzzleLogic Pte Ltd, Singapore 229594, Singapore \\
  [1em]
  \footnotesize \textbf{Corresponding author:} Fei Xia\texttt{<fei.xia@uci.edu>} \\
}

\begin{document}
\twocolumn[\begin{@twocolumnfalse}
\maketitle
\thispagestyle{empty}
\input{Sections/0_Abstract}
\vspace{0.35cm}
\end{@twocolumnfalse}]
\input{Sections/1_Introduction}
\input{Sections/2_Related_Works}
\input{Sections/3_Methods}

\input{Sections/4_Experiment}
\input{Sections/5_Discussion}

\input{Sections/6_Conclusion}
\bibliography{Sections/References}
\end{document}

%% file: Sections/0_Abstract.tex
\begin{abstract}
Histopathological whole slide images (WSIs) are central to cancer diagnosis, but their gigapixel scale and complex tissue heterogeneity make manual assessment time-consuming and prone to inconsistent interpretation. Deep learning has shown strong potential for WSI analysis, where multiple instance learning (MIL) has become a widely adopted framework for aggregating tile-level features into slide-level predictions; however, robust generalization remains challenging because slide-level labels provide weak supervision, diagnostically relevant regions are sparse, tissue composition varies across slides, and diagnostic evidence spans multiple magnifications. Rather than relying on architecture-specific redesign, pathology-aware augmentation offers a scalable way to improve the WSI training distribution itself, yet existing strategies often perturb tissue regions without preserving diagnostic relevance, slide-level context, or cross-scale structure.
Here, we propose SlideMix, a model-agnostic multimodal augmentation framework for MIL-based WSI analysis. SlideMix uses a retrieval-augmented vision-language model (VLM)-based Visual-Language Adaptive Region selector to identify diagnostically relevant regions and reduce weak-label noise, followed by In-place Tile Shuffling to mix feature embeddings within diagnostically meaningful tissue regions, while preserving slide-level context. A VLM-based soft-labeling module assigns supervision to the mixed samples, and a multi-factor, loss-driven, online Curriculum-Learning Feedback scheme adaptively controls shuffle granularity, feature similarity, and shuffle ratio to promote cross-scale representation learning. 
Across 11 WSI datasets comprising 20,523 slides, 8 diagnostic tasks, and 10 WSI backbones, SlideMix improves accuracy and generalization in most settings and compares favorably with established augmentation baselines, demonstrating a simple, plug-and-play route toward more robust and scalable digital pathology models.
The source code is available at https://github.com/Xia-Research-Lab/SlideMix.
\end{abstract}

\keywords{Whole slide image analysis \and multimodal augmentation \and curriculum learning \and multi-scale representation learning}

%% file: Sections/1_Introduction.tex
\section{Introduction}
\label{introduction}
Pathology diagnosis serves as the gold standard for cancer diagnosis, relying on microscopic image interpretation to ensure accurate disease identification and treatment planning~\cite{wsi}. In modern digital pathology, tissue biopsies are routinely digitized into gigapixel-scale Whole Slide Images (WSIs), which preserve rich spatial details and multi-scale tissue features. However, their massive size makes manual review both labor-intensive and dependent on highly specialized expertise, often leading to inconsistencies and variability in diagnostic outcomes~\cite{unpuzzle}.

\begin{figure}[t]
    \centering
    \includegraphics[width=\columnwidth]{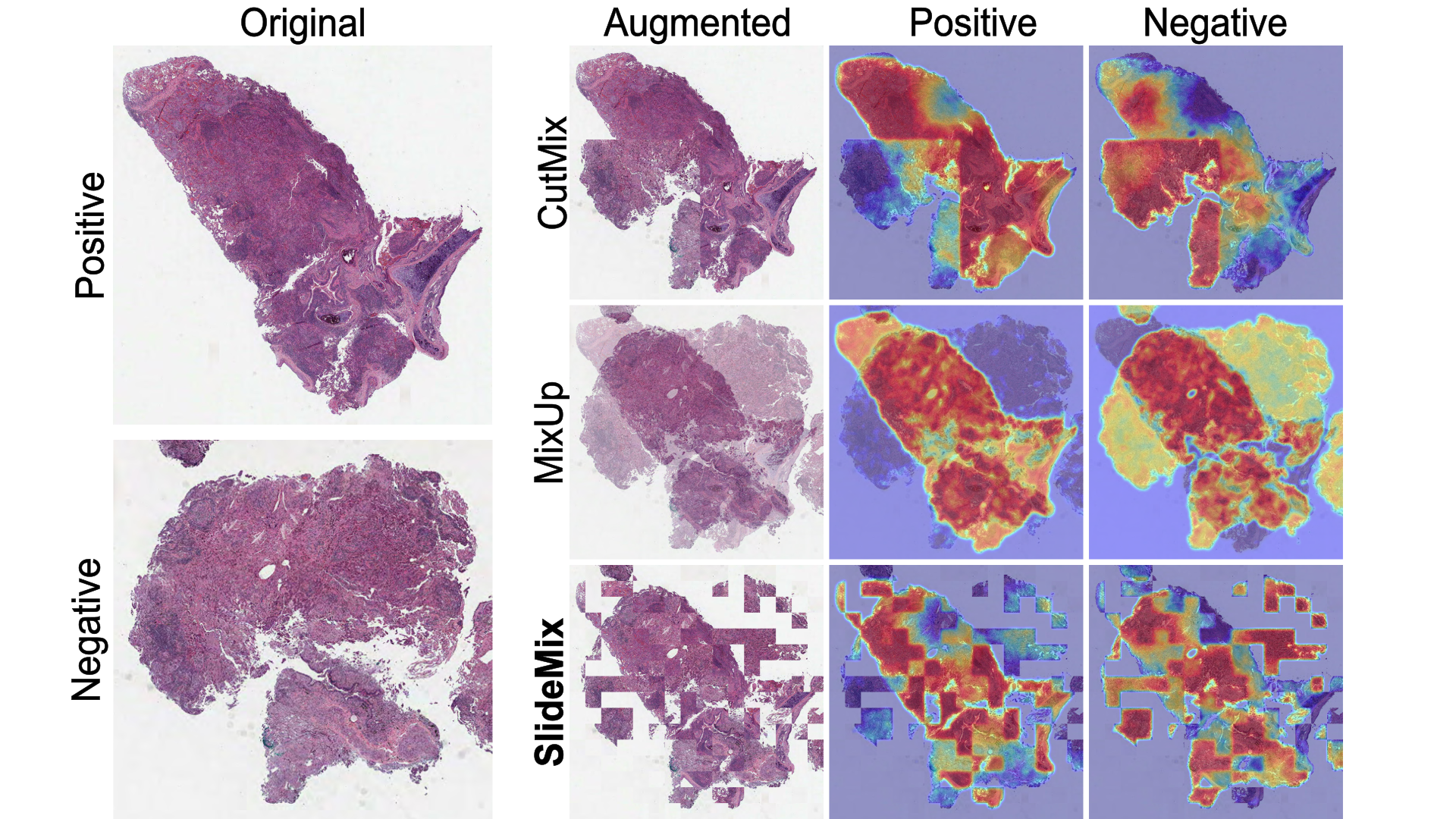}
    \caption{Visualization of an augmented example. Compared to other mixing-based methods, SlideMix accurately identifies tile and instance boundaries. It further distinguishes effectively between negative and positive samples.}
    \label{fig:heatmap}
\end{figure}

\begin{figure*}[t]
    \centering
    \includegraphics[width=\textwidth]{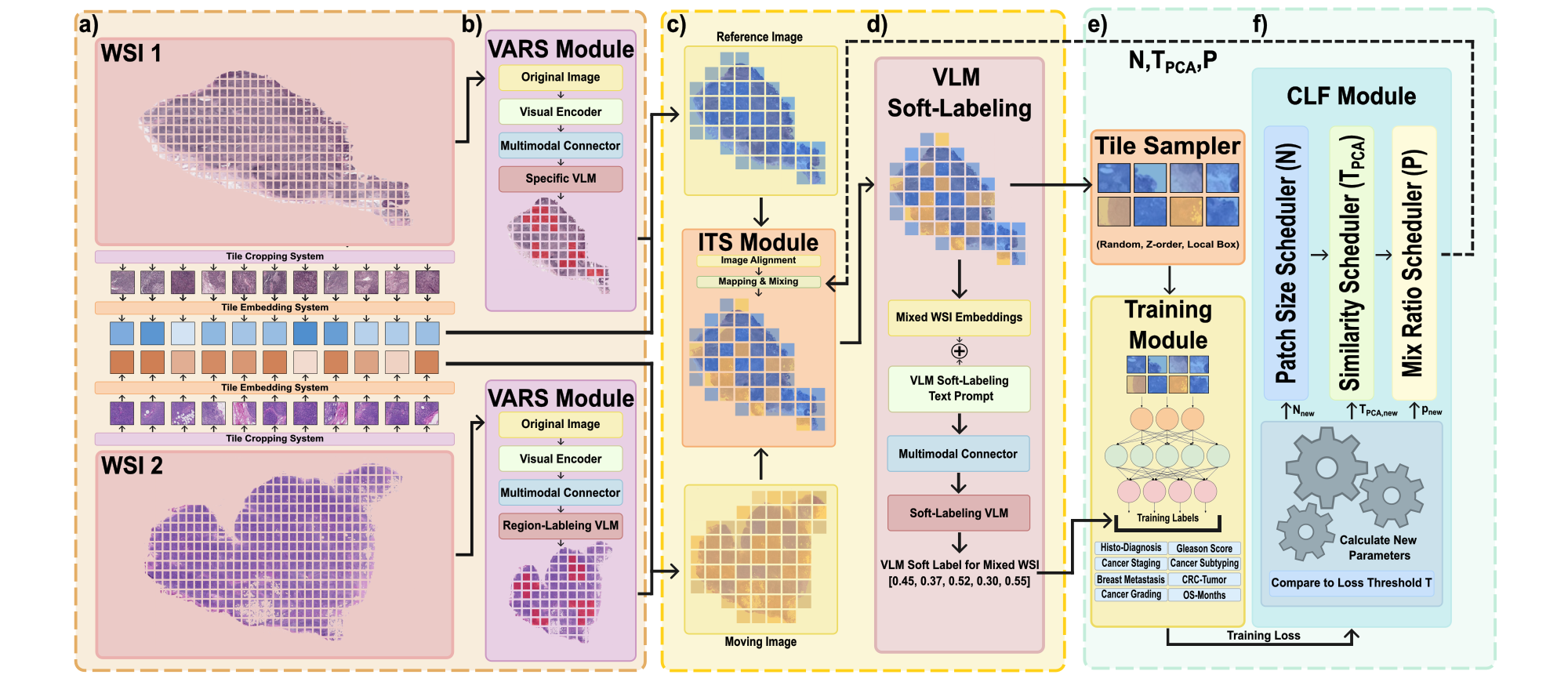}
    \caption{Overview of the MIL pipeline integrated with SlideMix. (a) Two WSIs from the same dataset are selected, tiled, and embedded into feature vectors for efficient computation. (b) The corresponding low-resolution WSIs are input into a fine-tuned VLM, which identifies label-relevant regions for mixing. (c) Coordinates of the selected candidate regions are passed to the data augmentation module, where the embedded tiles from stage one of MIL are shuffled in place according to three adjustable parameters $N$, $T_{PCA}$, and $P$ provided by the CLF, creating a new mixed sample. (d) The mixed sample is then processed by another separate VLM to generate a soft mixed label. (e) This new sample–label pair is fed into the tile sampler and subsequent training module. (f) Finally, the training loss is used to update the three aforementioned CLF parameters, enabling the ITS to automatically adjust the difficulty of the mixed samples throughout training.}
    \label{fig:main}
\end{figure*}

Deep learning techniques have been increasingly adopted to assist in WSI analysis by automatically identifying and categorizing critical histological patterns~\cite{cpia}. Recent methods employ Multiple Instance Learning (MIL), a two-stage framework designed to overcome GPU memory limits imposed by gigapixel-scale slides. Specifically, in the first stage of MIL, WSIs are divided into smaller tiles (e.g., $224\times224$ pixels from a $100,000\times80,000$-pixel WSI) and encoded into feature embeddings using a tile-level foundation model (e.g., UNI~\cite{uni}). In the second stage, the tile embeddings are aggregated into bags to produce slide-level predictions with a slide-level model (e.g., TransMIL~\cite{transmil})~\cite{cellmix}. Although practical, the inherent complexity and multi-scale nature of WSIs continue to challenge existing models, particularly in their ability to effectively integrate cross-scale information~\cite{cpia}. Three core challenges can thus be identified:

\noindent \textbf{1) Weak supervision from scarce fine-grained annotations}. 
Unlike natural images, only a small fraction of regions within a WSI (often $<1\%$) are label-relevant. This extreme imbalance between informative and non-informative regions greatly complicates the learning of discriminative features, as irrelevant tiles dominate the training process and weaken label-feature alignment.

\noindent \textbf{2) Spatial modeling impaired by local homogeneity and global heterogeneity}~\cite{cellmix}. 
Pathological structures exhibit strong local continuity but global diversity. This dual characteristic complicates spatial reasoning, as models must simultaneously preserve fine-grained local consistency while capturing long-range dependencies across the entire slide, a challenge that becomes increasingly pronounced with larger WSI dimensions.

\noindent \textbf{3) Limited multi-scale feature fusion capability}~\cite{chen2025pathrwkv}. 
WSIs contain features spanning multiple scales, from cellular to organ-level structures, and different diagnostic tasks emphasize distinct scales. For instance, tumor purity regression depends on global contextual understanding, while Epidermal Growth Factor Receptor (EGFR) mutation prediction demands precise modeling of cellular morphology.

To address these challenges, prior studies have focused on improving model architectures~\cite{abmil, chen2025pathrwkv} and developing data augmentation strategies~\cite{peng2025one, cong2022colour}. Architectural innovations can improve feature aggregation and spatial modeling, but they are often model-specific and do not directly address the weakly supervised and highly imbalanced training data presented to the model. In contrast, data augmentation can improve the training distribution itself and can be applied across different MIL architectures. However, existing augmentation methods often overlook key properties of pathological images (Fig. \ref{fig:heatmap}), including their multi-scale nature and severe label imbalance caused by large proportions of irrelevant regions. Moreover, most augmentation techniques remain static, limiting adaptability across diverse WSIs and tasks.

To this end, we propose SlideMix, a novel multimodal data augmentation framework for MIL-based WSI analysis (Fig. \ref{fig:main}). SlideMix integrates a Visual-Language Model (VLM)-guided region selector, an in-place tile shuffling mechanism, and an adaptive curriculum feedback loop. Together, these components address the aforementioned challenges through three key innovations:

\noindent 1) We propose a \textbf{VLM-based Adaptive Region (VAR) selector} that employs Retrieval-Augmented Generation (RAG) to retrieve domain-relevant knowledge and identify diagnostically significant ROIs at the lowest WSI scale, mitigating weak supervision effects.  

\noindent 2) We design an efficient \textbf{In-place Tile Shuffling (ITS) module} that mixes tile embeddings between WSIs using ROI coordinates from the VAR selector, balancing local homogeneity and global heterogeneity. A separate VLM generates soft labels from the mixed WSIs to create new training data.

\noindent 3) We introduce a \textbf{multi-factor Curriculum Learning Feedback (CLF) module} that adaptively adjusts the shuffle ratio, PCA similarity threshold, and shuffle granularity in the ITS module based on loss evaluation, enabling progressive cross-scale feature learning.

Experimental results across eight pathology tasks, ten state-of-the-art WSI models, and eleven WSI datasets comprising 20,523 WSIs demonstrate the robustness and adaptability of SlideMix. By providing new insights into multimodal feature regrouping, SlideMix improves model generalization and diagnostic accuracy in digital pathology applications.

%% file: Sections/2_Related_Works.tex
\section{Related Works}
\label{sec:related_works}
\subsection{Data Augmentation}
Data augmentation is a fundamental technique in deep learning, designed to enhance model generalization and robustness by generating modified or synthetic samples while preserving the information relevant to the learning task. For natural images, common methods include simple geometric and color transformations, such as flipping, rotating, and color jittering. However, these generic augmentations often fail to capture the unique and complex characteristics of pathological images. Consequently, domain-specific augmentation techniques have been developed. Early pathology-specific methods operated at the tile-level, including stain normalization~\cite{cong2022colour} and the generation of synthetic artifacts using GANs or latent-space models. Although more relevant to pathology, these methods overlook the broader spatial context and multi-scale features inherent in a WSI. More recent methods have shifted toward feature-level augmentation in the second stage of MIL to better model spatial relationships. For instance, Z-order sampling~\cite{peng2025one} processes tiles in Z-order to preserve the smallest spatial distance between the tiles, encouraging the model to learn spatial dependencies. However, such methods remain label-agnostic, applying uniformly across the entire WSI and inadvertently mixing large label-irrelevant regions with the few diagnostically critical ones. This introduces substantial noise and impairs the model's capacity to learn meaningful discriminative features.

\subsection{Visual Language Models}
Visual-Language Models (VLMs) have demonstrated remarkable capabilities in bridging vision and natural language, enabling complex reasoning that requires multimodal understanding~\cite{zhang2024vision}. These models are often pre-trained on large datasets of image-text pairs and excel at zero-shot generalization for tasks such as image classification, object detection, and visual question answering. In the medical domain, VLMs are increasingly applied to interpret complex medical imagery by leveraging associated textual information, such as clinical notes or pathology reports~\cite{lin2025taming}. Their ability to ground textual concepts within visual data makes them particularly promising for localizing ROIs in WSIs. However, most pathology applications of VLMs have focused on direct diagnosis or report generation. In contrast, our approach leverages the semantic recognition and reasoning capabilities of a VLM, not as a direct diagnostic tool, but as an intelligent guidance mechanism for identifying diagnostically relevant regions and directing pathology-aware data augmentation.

\subsection{Curriculum Learning}
Curriculum learning is a training strategy inspired by human cognition, in which models are presented with training examples in a structured order~\cite{bengio2009curriculum}. The model first learns simpler concepts and then progressively tackles more difficult ones, which facilitates faster convergence and improved performance. This principle is particularly suitable for WSI augmentation, because overly complex mixing early in training may obscure diagnostic patterns, whereas gradually increasing augmentation difficulty allows the model to first learn reliable pathological features before handling more challenging mixed samples. However, due to the complex and heterogeneous nature of WSIs, defining an effective curriculum in computational pathology is challenging. Previous WSI data augmentation methods applied fixed augmentation rules throughout training~\cite{peng2025one}, failing to adapt to the model's evolving learning state or to the varying complexity of different WSIs and diagnostic tasks.

%% file: Sections/3_Methods.tex
\section{Methods}
\subsection{Data Pre-processing and Tile Embedding} A WSI $S$ is first normalized to a unified resolution of 0.5 $\mu$m/px, ensuring invariance to raw scanner magnification and 
guaranteeing consistency across slides acquired from different scanners and institutions. The WSI is then partitioned into a non-overlapping grid of tiles $\{T_{\text{i,j}}\}$ of a 
corresponding size $T_{\text{size}}$ (Fig. \ref{fig:main}a). A two-stage filtering protocol is employed to ensure the quality of the selected regions. It first discards tiles with tissue coverage below a predefined threshold (e.g., $<50\%$ of the tile area), then removes tiles where the pixel variance falls below a quantitative cutoff (e.g., $Var(I)<0.01$, where $I$ represents the pixel intensity normalized to the $[0,1]$ range). These filtering steps exclude background-dominated and nearly uniform tiles that contain limited histological information, ensuring that downstream feature extraction and analysis focus on tissue-rich, visually informative regions. Following pre-processing, each filtered tile is embedded into a dense, semantic feature vector (Fig. \ref{fig:main}a) using a pathological foundation model (e.g., GigaPath \cite{gigapath}). This embedding process boosts both training efficiency and performance of the slide-level model (e.g., ABMIL \cite{abmil}).

\subsection{VLM-based Adaptive Region Selector}
\begin{figure}[t]
    \centering
    \includegraphics[width=\columnwidth]{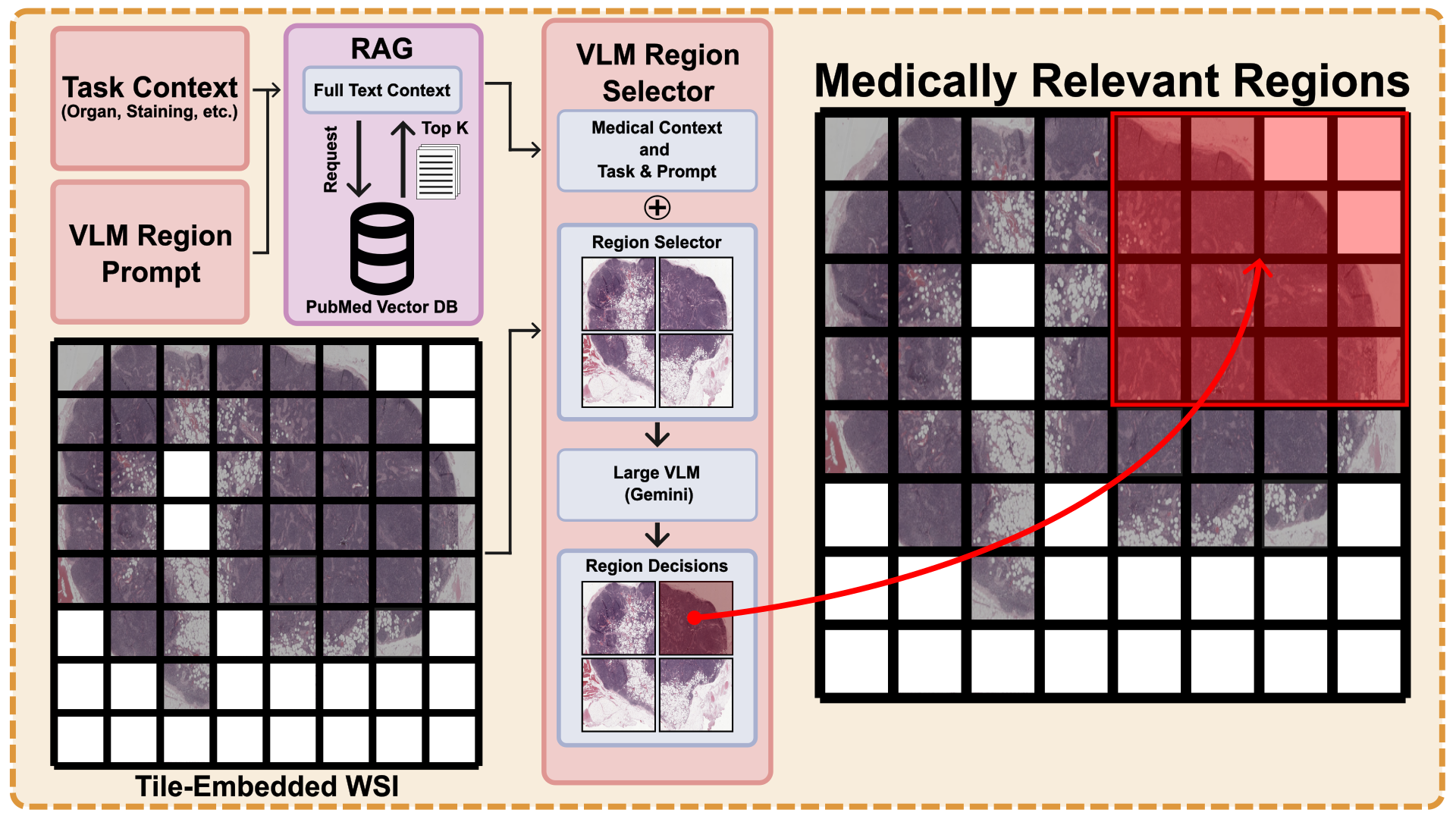}
    \caption{The proposed VAR selector: a) The task label $L$ and a visual-language model (VLM) region prompt $P$ are sent to a retrieval-augmented generation (RAG) system $R$ to retrieve the top-$K$ relevant medical data sources $D$ from PubMed $\mathcal{D}$. b) The WSI $S$ at lowest magnification is processed by the visual encoder $G_v$ of the RAG-augmented Gemini model $\text{Gem}_f$ to extract visual features $F$, which are partitioned into large regions. c) $F$, $L$, and $D$ are fused via the multimodal connector $G_m$ within $\text{Gem}_f$. d) Gemini 3.1 Pro iterates and predicts outcomes for each region, while the coordinate generator $G_p$ proposes a set of tile coordinates $C$ demarcating diagnostically relevant regions of interest (ROIs).}
    \label{fig:var}
\end{figure}

\subsubsection{Justification}
While a standalone VLM could serve as a region selector, its direct application to specialized medical domains risks generating hallucinated region proposals, as general-purpose VLMs are not inherently grounded in domain-specific clinical knowledge~\cite{xiong2024benchmarking}. To mitigate this, we augment the VLM with a Retrieval-Augmented Generation (RAG) system that queries PubMed in real-time, grounding each region proposal in established medical literature. RAG has been shown to consistently outperform standalone medical MLLMs on clinical reasoning tasks, with improvements extending to both frequent and rare pathological entities that are underrepresented in model training data~\cite{xiong2024benchmarking}. By coupling spatial reasoning from the VLM with retrieved biomedical evidence, VAR selector produces region proposals that are both visually informed and clinically contextualized, a property empirically validated in our ablation study 
(Tab.~\ref{tab:ablation1}).

\subsubsection{Algorithm}
To dynamically identify label-related regions in WSIs, we propose the VLM-based Adaptive Region (VAR) Selector (Fig.~\ref{fig:var}). Unlike conventional static selectors, the VAR Selector dynamically analyzes WSIs and proposes tile-level masks over diagnostically relevant regions, ensuring that the downstream ITS module shuffles only meaningful tissue rather than label-irrelevant background. This targeted shuffling reduces label noise and produces mixed samples that remain pathologically meaningful. The VAR selector integrates Gemini~\cite{team2023gemini} with the RAG system described above.

In a single augmentation process, two VAR selector instances handle two WSIs from the same dataset simultaneously (Fig.~\ref{fig:main}b). Each WSI $S$ is loaded at the lowest magnification level (highest mpp) for efficiency and passed into the visual encoder $G_v$ of the RAG-augmented Gemini model $\text{Gem}_f$ to extract visual features $F$:
\begin{equation}
F = G_v(S)
\end{equation}

\noindent Concurrently, the task context label $L$ (organ, staining method, etc.) and a VLM prompt $P$ (e.g., ``Identify key pathological regions in this WSI.'') are forwarded to $R$. The encoder $\phi$, implemented as BiomedCLIP~\cite{biomedclip} pretrained on 15 million biomedical image-text pairs from PubMed Central, tokenizes and encodes the concatenation of $L$ and $P$ into a dense query embedding $e_q = \phi([L\,;\,P]) \in \mathbb{R}^{512}$. Since all PubMed documents $d_i \in \mathcal{D}$ are pre-encoded offline into the same embedding space and indexed via FAISS~\cite{faiss}, the query $e_q$ can be efficiently matched against the entire database at augmentation time via approximate nearest-neighbor search, returning the top-$K$ most relevant literature sources $D$:
\begin{equation}
D = R_{\mathcal{D}}(L, P) = \underset{d_i \in \mathcal{D}}
{\text{top-}K}
\left(
\frac{\phi([L\,;\,P]) \cdot \phi(d_i)}
{\|\phi([L\,;\,P])\|\,\|\phi(d_i)\|}
\right)
\end{equation}

\noindent For example, given a survival prediction task with $L = \text{``Lymph Node, H\&E Staining''}$, $R$ retrieves literature on prognostic histological markers relevant to that tissue type, grounding region proposals in established clinical evidence. The multimodal connector $G_m$ of $\text{Gem}_f$ then concatenates $F$, $L$, and $D$ into a unified representation, from which the predictor $G_p$ generates a natural language description $T$ of diagnostically relevant regions (e.g., ``dense cellular clusters in the upper left and glandular structures in the lower center''):
\begin{equation}
T = G_p\bigl(G_m(F,\, L,\, D)\bigr)
\end{equation}

\noindent Since VLMs do not reliably produce precise numerical coordinates, we convert $T$ into tile-level coordinates via a grounding step. The VLM then iterates through the tiles and the similarity between each tile embedding $e_{ij}$ and the textual description $T$ is computed using $\phi$:
\begin{equation}
s_{ij} = \frac{\phi(T) \cdot e_{ij}}{\|\phi(T)\|\,\|e_{ij}\|}
\end{equation}

\noindent Tiles whose similarity score exceeds a threshold $\tau$ are retained to form the final coordinate set $C$:
\begin{equation}
C = \{(i,j) \mid s_{ij} > \tau\}
\end{equation}

\noindent The coordinates $C$ serve a dual purpose in the downstream pipeline: they constrain the ITS module to shuffle tiles exclusively within diagnostically relevant regions, while acting as a binary mask that excludes label-irrelevant background tiles from the mixing process entirely. This targeted selection ensures that augmented WSIs contain only meaningful tissue combinations, preventing background noise from diluting the supervision signal during training.

\subsection{Curriculum Learning Feedback Module}

\begin{figure}[t]
    \centering
    \includegraphics[width=1\columnwidth]{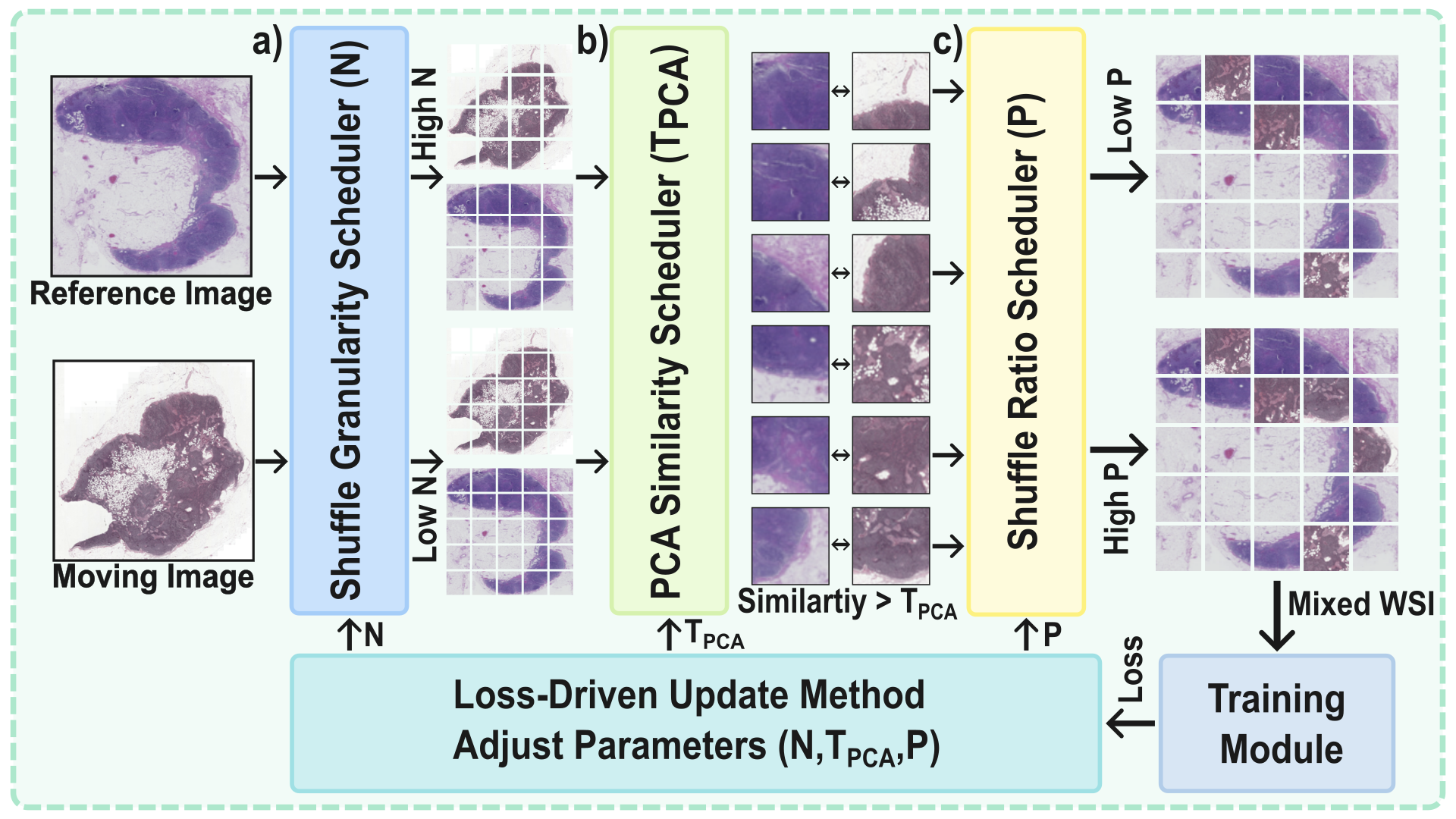}
    \caption{The proposed CLF module dynamically adjusts three schedulers based on training loss $l$ and threshold $T_{loss}$. 
    a) Shuffle Granularity ($N$) controls shuffling region scale. 
    b) PCA similarity threshold ($T_{PCA}$) governs feature similarity between tiles selected for shuffling. 
    c) Shuffle Ratio ($P$) sets the proportion of shuffled tiles. These updated parameters are then passed to the ITS module to configure the next augmentation.}
    \label{fig:clf}
\end{figure}

To enable gradual cross-scale feature learning, we implement the Curriculum Learning Feedback (CLF) module (Fig. \ref{fig:clf}). It dynamically adjusts the difficulty of the data augmentation in the downstream ITS module based on the model's performance. In CLF, the curriculum starts with simple augmentations and progressively increases the difficulty by three schedulers controlling shuffle granularity, PCA similarity threshold, and shuffle ratio. The difficulty is set by comparing the model's current training loss $l$ against a performance threshold $T_{loss}$.

\noindent \textbf{Shuffle Granularity ($N$)}: 
This scheduler determines the feature scale the model learns to recognize. It first groups tile embeddings into different sizes, denoted as $N \times N$. The curriculum starts with large $N$ (e.g., $16 \times 16$), a relatively easy task that lets the model learn the coarse-grained features (e.g., the boundaries between different tissues). It then gradually decreases $N$ in the sequence of $[N_0, N_1, ..., N_n]$, where $N_{i+1} < N_i$. In the end, the model will learn from the most fine-grained patterns (e.g., the features inside a cell).

\noindent \textbf{PCA Similarity Threshold ($T_{PCA}$)}: 
This scheduler determines the feature similarity the model learns to recognize. We use proximity in PCA space as a proxy for feature similarity. We define the distance $PCA_{\Delta}(e_i,e_j)$ between the principal components of two tiles $e_i$ and $e_j$ as:
\begin{equation}
PCA_{\Delta}(e_i,e_j)=|\mathrm{PCA}(e_i)-\mathrm{PCA}(e_j)|^2
\end{equation}

\noindent To ensure a consistent difficulty scale, we min-max normalize this distance to $[0,1]$ using the statistics ($PCA_{min}, PCA_{max}$) from the entire training set $\mathcal{D}$:

\begin{equation}
PCA'_{\Delta} = \frac{PCA_{\Delta} - PCA_{min}}{PCA_{max} - PCA_{min}}
\end{equation}

\noindent A threshold $T_{PCA} \in [0,1]$ constrains the shuffling: a tile pair can be shuffled only if $PCA'_{\Delta}(e_i, e_j) < T_{PCA}$. The curriculum starts with a relatively high $T_{PCA}$ (allowing dissimilar tiles to be shuffled), then gradually decreases the PCA similarity threshold, $[T_{PCA,0}, ..., T_{PCA,n}]$, where $T_{PCA, i+1} < T_{PCA, i}$. At the end, the model is forced to learn fine-grained differences, as only the most similar tiles are permitted to be shuffled.

\noindent \textbf{Shuffle Ratio ($P$)}: 
This scheduler determines the feature integrity the model learns to recognize. It controls the shuffle ratio, denoted as $P\in [0, 1]$. The curriculum starts with a low $P$, a relatively easy task that lets the model learn from relatively intact features (e.g., 90\% integrity). It then gradually increases $P$ in the sequence of $[P_0, P_1, ..., P_n]$, where $P_{i+1} > P_i$. In the end, the model will learn from the highly fragmented features.

The parameters of all three schedulers are updated at the end of each training epoch. We implement the loss-hold strategy as the update rule. Let $\boldsymbol{\Theta}_i = (P_i, N_i, T_{PCA,i})$ represent the curriculum parameters at epoch $i$. The parameters for the subsequent epoch $\boldsymbol{\Theta}_{i+1}$ are:

\begin{equation}
\boldsymbol{\Theta}_{i+1} = \begin{cases} (P_{i+1}, N_{i+1}, T_{PCA,i+1}) & \text{if } l < T_{loss} \\ \boldsymbol{\Theta}_i & \text{otherwise} \end{cases}
\end{equation}

\noindent This ensures difficulty only escalates after the model masters the current difficulty level.

\subsection{In-place Tile Shuffling Module}

\subsubsection{Justification}
Unlike natural image mixing, which is purely synthetic, SlideMix combines tissue regions from different WSIs in a manner that carries an inherent degree of biological plausibility. Tumor heterogeneity and diverse co-occurring tissue morphologies are intrinsic properties of real pathological slides \cite{marusyk2012intratumor}, meaning that the mixed samples produced by SlideMix reflect tissue configurations that plausibly occur in real diagnostic settings. Within a single WSI, regions of varying malignancy grade, stromal infiltration, and necrosis commonly co-exist in close spatial proximity \cite{marusyk2012intratumor}, and by recombining tissue regions across slides from the same diagnostic task, SlideMix simulates this natural heterogeneity in a principled way.

This biological grounding is further strengthened by a key property of MIL aggregators that is often overlooked: while most aggregators are invariant to tile order, they are not invariant to tile composition. ITS exploits this distinction by fundamentally altering the set of tiles within a bag, exposing the model to novel feature combinations it would never encounter through reordering alone. This forces the attention mechanism to learn diagnostic representations that are robust to varying tissue contexts, rather than overfitting to the specific tile compositions present in the training data.

\begin{figure}[t]
    \centering
    \includegraphics[width=\columnwidth]{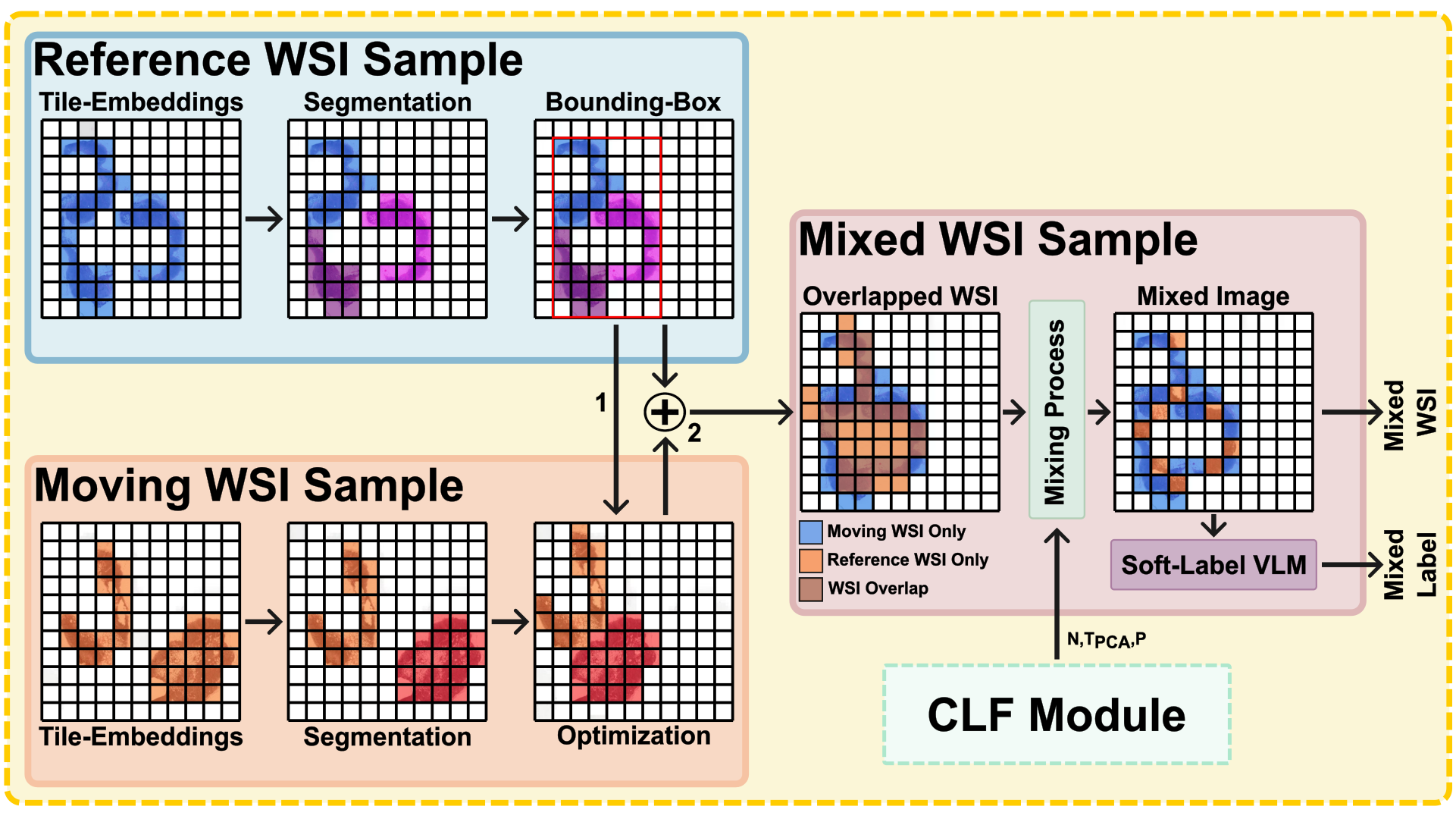}
    \caption{The ITS module operates as follows: a) a pair of embedded WSIs are provided and assigned as ‘reference’ and ‘moving’; b) each WSI is segmented into biological objects approximated by convex hulls; c) the reference WSI is enclosed within its minimal bounding box, and the moving WSI aligns to maximize overlap; d) overlapping regions are shuffled based on ITS configuration and CLF parameters; and e) soft labels for the mixed image are generated using a VLM, forming a new training sample.}
    \label{fig:its}
\end{figure}

\subsubsection{Algorithm}
The ITS module receives as input the tile embeddings from two WSIs, along with the coordinate sets $C_r$ and $C_m$ produced by their respective VAR selector instances for the reference WSI $S_r$ and moving WSI $S_m$. Only tile embeddings $e_{ij}$ whose coordinates fall within $C_r$ or $C_m$ are considered for shuffling; all remaining tiles are held fixed and carried through to the augmented WSI unchanged. This restriction ensures that the feature-level mixing performed by ITS operates exclusively over diagnostically relevant tissue, preserving the spatial context of label-irrelevant background regions.

The process starts with a pair of raw WSIs, designated as the static reference WSI ($S_r$) and the moving WSI ($S_m$).

To maximize overlap for meaningful WSI augmentation, we align the biological structures of $S_m$ to $S_r$ before shuffling. The spatial layout of a WSI $S$ is represented as a set of tile coordinates $T \subset \mathbb{Z}^2$. $T$ is partitioned into disjoint, connected components $\{z_1, z_2, ..., z_n\}$, where each $z_i$ corresponds to a distinct biological structure (e.g., a contiguous tissue section):

\begin{equation}
T = \bigcup_{i=1}^{n} z_i, \quad \text{where } z_i \cap z_j = \emptyset \text{ for } i \neq j
\end{equation}

\noindent Each component $z_i$ is treated as an independently movable object. We seek an optimal set of 2D translation vectors $\Theta = \{\boldsymbol{\theta}_1, ..., \boldsymbol{\theta}_n\}$ for $S_m$'s components ($z_{m,i}$) to maximize the spatial overlap with $S_r$'s coordinates $T_r$:

\begin{equation}
\Theta^* = \underset{\Theta}{\arg\max} \left| \left( \bigcup_{i=1}^n (z_{m,i} + \boldsymbol{\theta}_i) \right) \cap T_r \right|
\end{equation}

\noindent Directly optimizing this objective on millions of tiles is computationally prohibitive. To create a tractable problem, we approximate each tissue component $z_i$ with its convex hull $\operatorname{Conv}(z_i)$. A standard optimizer then solves for the translations $\Theta^*$ that align the convex hulls. Once found, these optimal offsets are applied to the original tile coordinates of $S_m$ to produce an aligned WSI $S_a$.

The shuffling process occurs within the aligned, overlapping region $T_r \cap T_m$ in $S_a$. The coordinates of tile embeddings from $S_m$ and $S_r$ are randomly shuffled according to the configurations ($P, N, T_{PCA}$) from ITS. This shuffled WSI $S_s$ contains a semantically coherent fusion of tissue structures from both $S_r$ and $S_m$.

We then integrate CONCH \cite{lu2024visual}, a pathological VLM, with our RAG system as a soft-labeler to generate the corresponding soft labels $L_s$. Similar to Gemini in the VAR selector, CONCH sends the labels and prompts to the PubMed database and retrieves medical knowledge for assistance. The CONCH VLM uses a different embedding system, so the mixed image is re-encoded with CONCH embeddings only for soft labeling.

\subsection{Slide-level Feature Modeling}
The augmented WSI $S_s$ and label $L_s$ are passed to the slide-level backbone (e.g., TransMIL \cite{transmil}) to generate the final slide-level predictions for different downstream tasks, following the conventional MIL pipeline. At the end of the epoch, the validation loss is sent back to CLF to adjust the parameters for the next augmentation.

\begin{figure}[H]
    \centering
    \includegraphics[width=1\columnwidth]{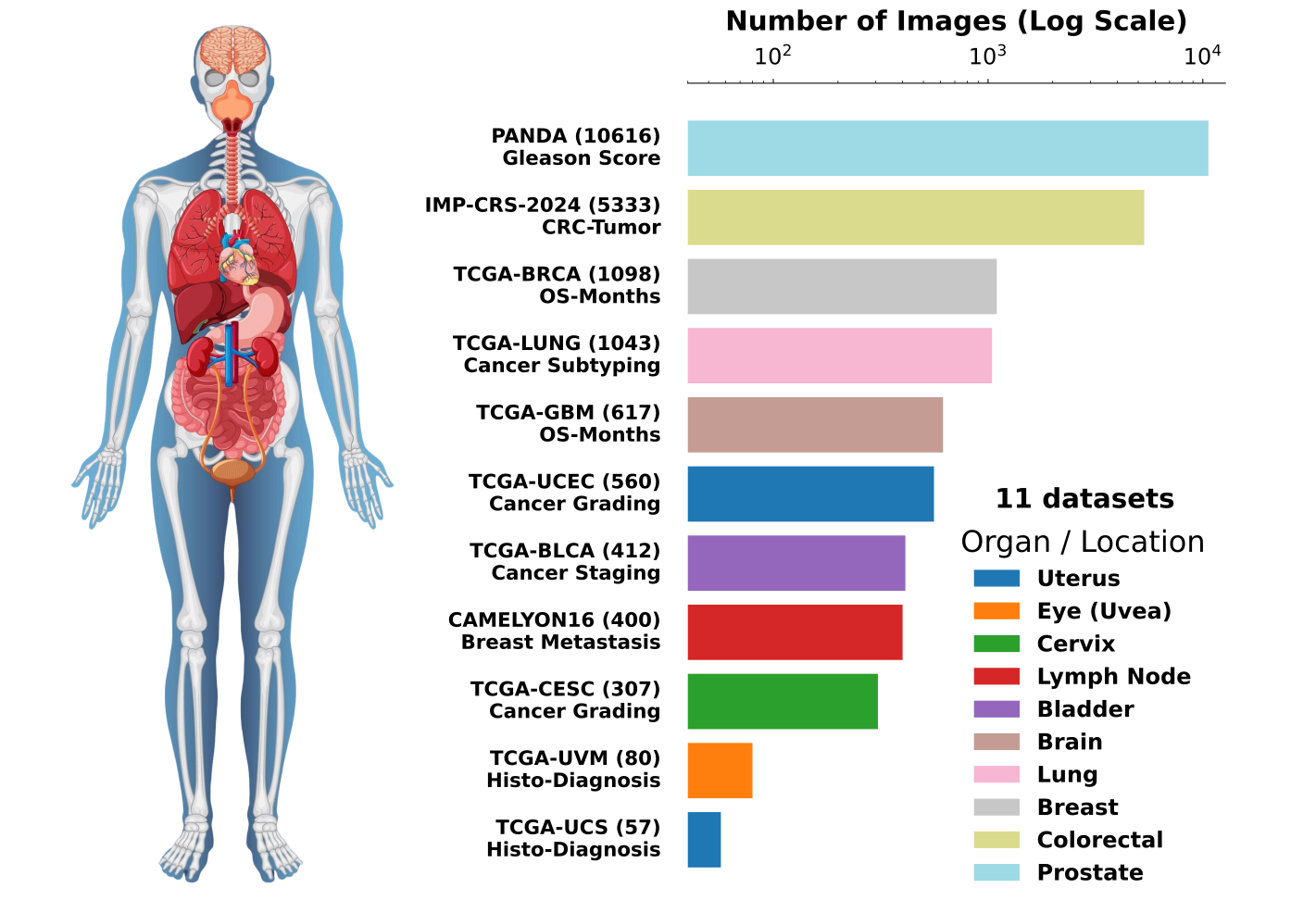}
    \caption{Summary of implemented datasets and tasks.}
    \label{fig:datasets}
\end{figure}

%% file: Sections/4_Experiment.tex
\section{Experiment}

\begin{table*}
\centering
\caption{Benchmarking SlideMix against SOTA augmentation methods with ABMIL as the baseline model. Results are reported as mean $\pm$ standard deviation over five runs. Within numeric result cells, bold denotes the highest mean in each dataset column; all ties at the reported precision are bolded.}
\label{tab:comparison1}
\resizebox{\textwidth}{!}{
\begin{tabular}{lccccccccccc}
\toprule
\multirow{2}{*}{\textbf{Method}} & \textbf{PANDA} & \textbf{CAMELYON16} & \textbf{IMP-CRS-2024} & \textbf{TCGA-Lung} & \textbf{TCGA-BLCA} & \textbf{TCGA-UCEC} & \textbf{TCGA-CESC} & \textbf{TCGA-UCS} & \textbf{TCGA-UVM} & \textbf{TCGA-BRCA} & \textbf{TCGA-GBM} \\
\cmidrule{2-12}
& \textbf{Acc. [\%]} & \textbf{Acc. [\%]} & \textbf{Acc. [\%]} & \textbf{Acc. [\%]} & \textbf{Acc. [\%]} & \textbf{Acc. [\%]} & \textbf{Acc. [\%]} & \textbf{Acc. [\%]} & \textbf{Acc. [\%]} & \textbf{Corr.} & \textbf{Corr.} \\
\midrule
\textbf{Baseline}  & 75.9$\pm$1.1 & 93.7$\pm$0.9 & 95.2$\pm$0.6 & 71.6$\pm$1.6 & 48.4$\pm$2.1 & 67.5$\pm$1.5 & 54.4$\pm$1.9 & 58.3$\pm$3.8 & 62.5$\pm$3.5 & 0.449$\pm$0.032 & 0.614$\pm$0.026 \\
\textbf{MixUp}     & 76.3$\pm$1.0 & 94.1$\pm$0.8 & 95.4$\pm$0.5 & 72.1$\pm$1.5 & 49.2$\pm$2.0 & 68.1$\pm$1.4 & 54.8$\pm$1.8 & 59.1$\pm$3.6 & 63.2$\pm$3.3 & 0.461$\pm$0.030 & 0.625$\pm$0.025 \\
\textbf{CutMix}    & 76.8$\pm$1.0 & 94.5$\pm$0.8 & 95.6$\pm$0.5 & 72.5$\pm$1.5 & 49.7$\pm$2.0 & 68.6$\pm$1.4 & 55.2$\pm$1.8 & 59.8$\pm$3.5 & 63.8$\pm$3.2 & 0.472$\pm$0.029 & 0.628$\pm$0.025 \\
\textbf{CutOut}    & 75.2$\pm$1.2 & 93.1$\pm$1.0 & 94.8$\pm$0.7 & 71.2$\pm$1.7 & 47.8$\pm$2.2 & 66.9$\pm$1.6 & 53.9$\pm$2.0 & 57.5$\pm$3.9 & 61.8$\pm$3.6 & 0.438$\pm$0.033 & 0.608$\pm$0.027 \\
\textbf{ResizeMix} & 77.1$\pm$0.9 & 94.8$\pm$0.7 & \textbf{95.8$\pm$0.5} & 72.9$\pm$1.4 & 50.1$\pm$1.9 & 69.1$\pm$1.3 & \textbf{55.6$\pm$1.7} & 60.3$\pm$3.4 & 64.2$\pm$3.1 & 0.478$\pm$0.028 & 0.632$\pm$0.024 \\
\textbf{PuzzleMix} & 76.6$\pm$1.0 & 94.3$\pm$0.8 & 95.5$\pm$0.5 & 72.3$\pm$1.5 & 49.5$\pm$2.0 & 68.4$\pm$1.4 & 55.0$\pm$1.8 & 59.5$\pm$3.5 & 63.5$\pm$3.2 & 0.467$\pm$0.029 & 0.627$\pm$0.025 \\
\textbf{SlideMix}  & \textbf{77.2$\pm$0.8} & \textbf{94.9$\pm$0.6} & \textbf{95.8$\pm$0.4} & \textbf{73.8$\pm$1.2} & \textbf{51.6$\pm$1.7} & \textbf{69.8$\pm$1.1} & 55.3$\pm$1.5 & \textbf{66.7$\pm$3.0} & \textbf{68.8$\pm$2.7} & \textbf{0.521$\pm$0.024} & \textbf{0.637$\pm$0.021} \\
\bottomrule
\end{tabular}}
\end{table*}

\begin{table*}
\centering
\caption{Impact of SlideMix on SOTA MIL backbone performance. Results are reported as mean $\pm$ standard deviation over five runs. Within each Base/Ours pair, bold denotes the higher mean; when the means are equal at the reported precision, both values are bolded.}
\label{tab:comparison2}
 
\resizebox{\textwidth}{!}{
\begin{tabular}{lcccccccccccc}
\toprule
\multirow{3}{*}{\textbf{Method}} & \multicolumn{2}{c}{\textbf{PANDA}} & \multicolumn{2}{c}{\textbf{CAMELYON16}} & \multicolumn{2}{c}{\textbf{IMP-CRS-2024}} & \multicolumn{2}{c}{\textbf{TCGA-Lung}} & \multicolumn{2}{c}{\textbf{TCGA-BLCA}} & \multicolumn{2}{c}{\textbf{TCGA-UCEC}} \\
\cmidrule(lr){2-3} \cmidrule(lr){4-5} \cmidrule(lr){6-7} \cmidrule(lr){8-9} \cmidrule(lr){10-11} \cmidrule(lr){12-13}
& \multicolumn{2}{c}{\textbf{Acc. [\%]}} & \multicolumn{2}{c}{\textbf{Acc. [\%]}} & \multicolumn{2}{c}{\textbf{Acc. [\%]}} & \multicolumn{2}{c}{\textbf{Acc. [\%]}} & \multicolumn{2}{c}{\textbf{Acc. [\%]}} & \multicolumn{2}{c}{\textbf{Acc. [\%]}} \\
\cmidrule(lr){2-3} \cmidrule(lr){4-5} \cmidrule(lr){6-7} \cmidrule(lr){8-9} \cmidrule(lr){10-11} \cmidrule(lr){12-13}
& Base & Ours & Base & Ours & Base & Ours & Base & Ours & Base & Ours & Base & Ours \\
\midrule
\textbf{SlideAve}
& 66.7$\pm$1.8 & \textbf{68.1$\pm$1.5}
& 69.6$\pm$1.9 & \textbf{71.2$\pm$1.6}
& 92.3$\pm$0.9 & \textbf{93.2$\pm$0.7}
& \textbf{75.1$\pm$1.3} & 74.6$\pm$1.3
& 51.1$\pm$2.1 & \textbf{52.2$\pm$1.8}
& 68.3$\pm$1.6 & \textbf{69.6$\pm$1.3} \\
\textbf{SlideMax}
& 62.5$\pm$1.9 & \textbf{64.2$\pm$1.6}
& \textbf{95.3$\pm$0.7} & 94.6$\pm$0.7
& 93.1$\pm$0.9 & \textbf{94.1$\pm$0.7}
& \textbf{75.7$\pm$1.3} & 75.1$\pm$1.3
& 53.8$\pm$1.9 & \textbf{54.9$\pm$1.6}
& 69.2$\pm$1.5 & \textbf{70.4$\pm$1.3} \\
\textbf{ABMIL}
& 75.9$\pm$1.1 & \textbf{77.2$\pm$0.8}
& 93.7$\pm$0.9 & \textbf{94.9$\pm$0.6}
& 95.2$\pm$0.6 & \textbf{95.8$\pm$0.4}
& 71.6$\pm$1.6 & \textbf{73.8$\pm$1.2}
& 48.4$\pm$2.1 & \textbf{51.6$\pm$1.7}
& 67.5$\pm$1.5 & \textbf{69.8$\pm$1.1} \\
\textbf{CLAM}
& 76.0$\pm$1.1 & \textbf{76.8$\pm$0.9}
& 73.4$\pm$1.8 & \textbf{74.7$\pm$1.5}
& 92.9$\pm$0.9 & \textbf{93.7$\pm$0.7}
& 72.2$\pm$1.5 & \textbf{73.2$\pm$1.2}
& \textbf{50.0$\pm$2.0} & 49.5$\pm$1.8
& 64.2$\pm$1.8 & \textbf{65.8$\pm$1.5} \\
\textbf{DSMIL}
& \textbf{78.5$\pm$0.8} & 78.1$\pm$0.8
& 86.1$\pm$1.4 & \textbf{87.3$\pm$1.1}
& 94.6$\pm$0.7 & \textbf{95.4$\pm$0.5}
& 73.4$\pm$1.3 & \textbf{74.3$\pm$1.1}
& \textbf{56.8$\pm$1.8} & 55.9$\pm$1.7
& \textbf{71.7$\pm$1.3} & 70.8$\pm$1.2 \\
\textbf{TransMIL}
& \textbf{76.2$\pm$0.9} & 75.8$\pm$0.9
& 93.7$\pm$0.8 & \textbf{94.3$\pm$0.7}
& 95.2$\pm$0.6 & \textbf{95.9$\pm$0.4}
& 73.4$\pm$1.3 & \textbf{74.1$\pm$1.1}
& 46.6$\pm$2.3 & \textbf{48.4$\pm$1.9}
& \textbf{69.2$\pm$1.4} & 68.3$\pm$1.4 \\
\textbf{SETMIL}
& \textbf{78.2$\pm$0.8} & 78.0$\pm$0.8
& \textbf{94.1$\pm$0.7} & 93.8$\pm$0.7
& \textbf{95.6$\pm$0.5} & 95.5$\pm$0.5
& 75.8$\pm$1.2 & \textbf{76.0$\pm$1.0}
& \textbf{57.3$\pm$1.7} & 57.1$\pm$1.6
& 70.5$\pm$1.3 & \textbf{70.6$\pm$1.2} \\
\textbf{DTFD-MIL}
& \textbf{78.9$\pm$0.8} & 78.7$\pm$0.8
& \textbf{94.8$\pm$0.6} & 94.5$\pm$0.7
& 95.9$\pm$0.4 & \textbf{96.0$\pm$0.4}
& \textbf{85.2$\pm$0.9} & 85.0$\pm$0.9
& 58.1$\pm$1.6 & \textbf{58.3$\pm$1.5}
& \textbf{71.2$\pm$1.3} & 71.0$\pm$1.3 \\
\textbf{GigaPath}
& 77.9$\pm$0.9 & \textbf{78.5$\pm$0.8}
& 84.5$\pm$1.4 & \textbf{85.8$\pm$1.2}
& 94.9$\pm$0.6 & \textbf{95.1$\pm$0.5}
& 75.1$\pm$1.3 & \textbf{75.7$\pm$1.1}
& 52.2$\pm$1.9 & \textbf{53.8$\pm$1.7}
& 68.3$\pm$1.5 & \textbf{69.2$\pm$1.3} \\
\textbf{MambaMIL}
& \textbf{78.6$\pm$0.8} & 78.4$\pm$0.8
& 94.5$\pm$0.7 & \textbf{94.7$\pm$0.6}
& \textbf{95.7$\pm$0.5} & 95.6$\pm$0.5
& 76.3$\pm$1.1 & \textbf{76.5$\pm$1.0}
& \textbf{58.9$\pm$1.6} & 58.7$\pm$1.5
& 70.8$\pm$1.3 & \textbf{71.0$\pm$1.2} \\
\bottomrule
\end{tabular}}
\resizebox{\textwidth}{!}{
\begin{tabular}{lcccccccccc}
\toprule
\multirow{3}{*}{\textbf{Method}} & \multicolumn{2}{c}{\textbf{TCGA-CESC}} & \multicolumn{2}{c}{\textbf{TCGA-UCS}} & \multicolumn{2}{c}{\textbf{TCGA-UVM}} & \multicolumn{2}{c}{\textbf{TCGA-BRCA}} & \multicolumn{2}{c}{\textbf{TCGA-GBM}} \\
\cmidrule(lr){2-3} \cmidrule(lr){4-5} \cmidrule(lr){6-7} \cmidrule(lr){8-9} \cmidrule(lr){10-11}
& \multicolumn{2}{c}{\textbf{Acc. [\%]}} & \multicolumn{2}{c}{\textbf{Acc. [\%]}} & \multicolumn{2}{c}{\textbf{Acc. [\%]}} & \multicolumn{2}{c}{\textbf{Corr.}} & \multicolumn{2}{c}{\textbf{Corr.}} \\
\cmidrule(lr){2-3} \cmidrule(lr){4-5} \cmidrule(lr){6-7} \cmidrule(lr){8-9} \cmidrule(lr){10-11}
& Base & Ours & Base & Ours & Base & Ours & Base & Ours & Base & Ours \\
\midrule
\textbf{SlideAve}
& \textbf{52.6$\pm$2.3} & \textbf{52.6$\pm$2.0}
& 58.3$\pm$3.8 & \textbf{62.5$\pm$3.2}
& \textbf{75.0$\pm$2.9} & 71.9$\pm$2.7
& 0.53$\pm$0.03 & \textbf{0.54$\pm$0.03}
& 0.54$\pm$0.03 & \textbf{0.56$\pm$0.03} \\
\textbf{SlideMax}
& 47.4$\pm$2.5 & \textbf{49.1$\pm$2.2}
& 41.7$\pm$4.2 & \textbf{45.8$\pm$3.7}
& 31.2$\pm$4.6 & \textbf{37.5$\pm$4.0}
& \textbf{0.52$\pm$0.03} & 0.51$\pm$0.03
& 0.46$\pm$0.03 & \textbf{0.49$\pm$0.03} \\
\textbf{ABMIL}
& 54.4$\pm$1.9 & \textbf{55.3$\pm$1.5}
& 58.3$\pm$3.8 & \textbf{66.7$\pm$3.0}
& 62.5$\pm$3.5 & \textbf{68.8$\pm$2.7}
& 0.45$\pm$0.03 & \textbf{0.52$\pm$0.02}
& 0.61$\pm$0.03 & \textbf{0.64$\pm$0.02} \\
\textbf{CLAM}
& \textbf{54.4$\pm$1.9} & 52.6$\pm$1.8
& \textbf{58.3$\pm$3.8} & \textbf{58.3$\pm$3.4}
& 68.8$\pm$3.2 & \textbf{71.9$\pm$2.7}
& 0.54$\pm$0.03 & \textbf{0.55$\pm$0.02}
& 0.29$\pm$0.04 & \textbf{0.31$\pm$0.03} \\
\textbf{DSMIL}
& 50.9$\pm$2.1 & \textbf{52.6$\pm$1.8}
& 58.3$\pm$3.7 & \textbf{63.3$\pm$3.1}
& 62.5$\pm$3.5 & \textbf{65.6$\pm$2.9}
& 0.53$\pm$0.03 & \textbf{0.55$\pm$0.02}
& 0.51$\pm$0.03 & \textbf{0.53$\pm$0.03} \\
\textbf{TransMIL}
& 50.9$\pm$2.1 & \textbf{52.6$\pm$1.8}
& 66.7$\pm$3.1 & \textbf{70.8$\pm$2.6}
& 56.2$\pm$3.5 & \textbf{59.4$\pm$3.0}
& 0.50$\pm$0.03 & \textbf{0.52$\pm$0.03}
& \textbf{0.65$\pm$0.02} & 0.64$\pm$0.02 \\
\textbf{SETMIL}
& 55.1$\pm$1.8 & \textbf{55.2$\pm$1.6}
& 68.9$\pm$2.9 & \textbf{69.1$\pm$2.7}
& 71.8$\pm$2.7 & \textbf{72.0$\pm$2.5}
& \textbf{0.56$\pm$0.02} & \textbf{0.56$\pm$0.02}
& \textbf{0.63$\pm$0.02} & \textbf{0.63$\pm$0.02} \\
\textbf{DTFD-MIL}
& \textbf{55.8$\pm$1.7} & 55.5$\pm$1.6
& \textbf{70.3$\pm$2.6} & 70.1$\pm$2.5
& 72.5$\pm$2.6 & \textbf{72.6$\pm$2.4}
& \textbf{0.57$\pm$0.02} & \textbf{0.57$\pm$0.02}
& \textbf{0.64$\pm$0.02} & \textbf{0.64$\pm$0.02} \\
\textbf{GigaPath}
& 50.9$\pm$2.1 & \textbf{52.6$\pm$1.9}
& 50.0$\pm$4.0 & \textbf{54.2$\pm$3.4}
& 56.2$\pm$3.6 & \textbf{59.4$\pm$3.0}
& \textbf{0.64$\pm$0.02} & 0.63$\pm$0.02
& 0.62$\pm$0.02 & \textbf{0.63$\pm$0.02} \\
\textbf{MambaMIL}
& \textbf{54.7$\pm$1.8} & 54.5$\pm$1.7
& 69.5$\pm$2.8 & \textbf{69.8$\pm$2.6}
& 73.1$\pm$2.6 & \textbf{73.3$\pm$2.4}
& \textbf{0.57$\pm$0.02} & \textbf{0.57$\pm$0.02}
& \textbf{0.63$\pm$0.02} & \textbf{0.63$\pm$0.02} \\
\bottomrule
\end{tabular}}
\end{table*}

\subsection{Datasets and Downstream Tasks}
To demonstrate its effectiveness and generalizability, we evaluated SlideMix on 11 datasets across 8 downstream tasks covering diverse diagnostic scenarios (Fig.~\ref{fig:datasets}). The \textbf{PANDA}~\cite{panda} dataset with the \textbf{Gleason Score} task assesses prostate cancer aggressiveness, while \textbf{CAMELYON16}~\cite{camelyon16} with the \textbf{Breast Metastasis} task classifies lymph nodes as normal or tumorous. \textbf{IMP-CRS-2024}~\cite{impcrs2024} with the \textbf{CRS-Tumor} task identifies tumor tissues in colorectal images. The \textbf{TCGA}~\cite{tcga} datasets cover multiple cancer types and tasks: \textbf{TCGA-Lung} performs \textbf{Cancer Subtyping}, distinguishing lung cancer variants; \textbf{TCGA-BLCA} performs \textbf{Cancer Staging}, assessing bladder cancer progression; \textbf{TCGA-UCEC} and \textbf{TCGA-CESC} perform \textbf{Cancer Grading} on uterine and cervical tissues, respectively, determining tumor differentiation levels; \textbf{TCGA-UCS} and \textbf{TCGA-UVM} perform \textbf{Histological Diagnosis}, classifying uterine and uveal tissue subtypes based on morphology; and \textbf{TCGA-BRCA} and \textbf{TCGA-GBM} perform the \textbf{OS-Months} task, predicting patient survival time in months from breast and brain tissue morphology, respectively.

\subsection{Implementation Details}
The comparison in Tab.~\ref{tab:comparison1} uses ABMIL~\cite{abmil} as the slide-level backbone, while all backbones in Tab.~\ref{tab:comparison2} are initialized from scratch without pretraining. Models were trained for 100 epochs (20 for warmup and 80 for main training) using the Adam optimizer with an initial learning rate of $1 \times 10^{-4}$ and a cosine decay schedule annealing to $1 \times 10^{-6}$ ($1\%$ of the initial value). The UnPuzzle~\cite{unpuzzle} framework was used for pre-processing, with each WSI divided into $224 \times 224$ tiles represented by GigaPath~\cite{gigapath} feature embeddings for prediction, and into $512 \times 512$ tiles for soft-labeling using CONCH~\cite{lu2024visual}. To minimize sampling variance, particularly for small datasets, five independent test inferences with different random tile samplings were performed per scenario, and predictions were aggregated. A batch size of 4 was maintained for all experiments, conducted on an NVIDIA H100 GPU using Python~3.10.16, PyTorch~2.4.0, and CUDA~12.4. Across Tabs.~\ref{tab:comparison1}--\ref{tab:ablation3}, only boldface is used to emphasize numeric results; underlining and color highlighting are not used. Unless a caption states a metric-specific direction, bold denotes the best mean within the relevant comparison group, and all ties at the reported precision are bolded.

\subsection{Comparison with SOTA Methods}
We benchmarked SlideMix against five augmentation strategies, including \textbf{CutMix}~\cite{cutmix}, \textbf{CutOut}~\cite{cutout}, \textbf{MixUp}~\cite{mixup}, \textbf{ResizeMix}~\cite{resizemix}, and \textbf{PuzzleMix}~\cite{puzzlemix}. However, these methods operate on raw tiles or leverage saliency maps, making them directly incompatible with the MIL framework. To enable a fair comparison, we applied each augmentation at the raw image level and then computed tile embeddings from the resulting augmented images. For faithful implementation, we used the available original code repositories for each baseline method. The baseline model was trained without any data augmentation.

Tab.~\ref{tab:comparison1} shows that SlideMix performs strongly across the evaluated datasets. SlideMix has the highest mean in nine of the 11 dataset columns and ties ResizeMix on IMP-CRS-2024 (95.8\%), placing first or tied first in 10 of 11 columns; on TCGA-CESC, ResizeMix is slightly higher than SlideMix (55.6\% vs. 55.3\%). On CAMELYON16, SlideMix reaches 94.9\%, which is 1.2 percentage points above the baseline and 0.4 points above CutMix. SlideMix also exceeds the next-best method on PANDA by 0.1 points and on TCGA-Lung by 0.9 points. Thus, the direct comparison demonstrates a consistent overall advantage across diverse pathology tasks.

To evaluate the generalizability of SlideMix, we compared two pooling baselines, \textbf{SlideAve} and \textbf{SlideMax}, and eight MIL backbones: \textbf{ABMIL} \cite{abmil}, \textbf{DSMIL} \cite{dsmil}, \textbf{CLAM} \cite{clam}, \textbf{TransMIL} \cite{transmil}, \textbf{SETMIL} \cite{setmil}, \textbf{DTFD-MIL} \cite{dtfd}, \textbf{GigaPath} \cite{gigapath}, and \textbf{MambaMIL} \cite{mambamil}. Across the 110 Base/Ours comparisons in Tab.~\ref{tab:comparison2}, SlideMix improves 74, ties 8, and decreases 28 at the reported precision. ABMIL improves on all 11 datasets and GigaPath improves on ten, demonstrating strong compatibility across distinct backbones. The largest decreases occur for SlideAve on TCGA-UVM and CLAM on TCGA-CESC, identifying clear targets for backbone-specific calibration. Overall, these results demonstrate broad transfer across architectures without requiring backbone redesign.

%% file: Sections/5_Discussion.tex
\section{Discussion}
\subsection{Visualization Analysis}

\begin{figure*}[t]
    \centering
    \includegraphics[width=1\textwidth]{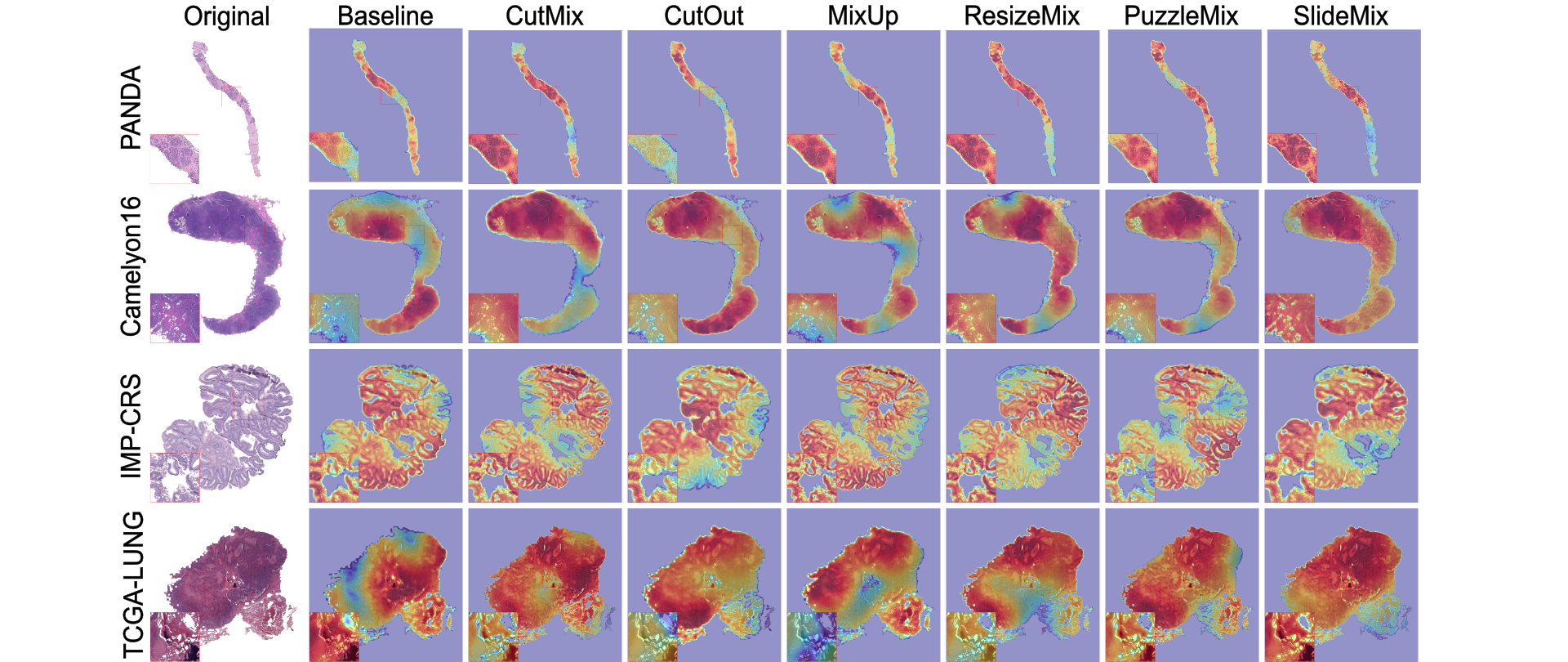}
    \caption{Grad-CAM visualization for ABMIL trained with different augmentation methods; the baseline uses no augmentation. All images are normalized on the same scale. Warmer colors indicate stronger activation. In the CAMELYON16 example, SlideMix shows more spatially concentrated activation than CutMix. The corresponding ABMIL accuracies are 94.9\% for SlideMix, 94.5\% for CutMix, and 93.7\% for the baseline (Tab.~\ref{tab:comparison1}).}
    \label{fig:gradcam}
\end{figure*}

To examine how augmentation changes model attention, we visualize Class Activation Maps using Grad-CAM \cite{gradcam}. All attention values were normalized on the same scale. In the representative PANDA and IMP-CRS-2024 examples, the SlideMix maps appear more spatially concentrated or boundary-aligned than several baselines. On CAMELYON16, SlideMix also suppresses diffuse activation relative to CutMix and yields a more spatially focused response. Because these examples do not include pixel-level ground-truth lesion annotations, the CAMs provide qualitative evidence that complements the quantitative results. We therefore refer readers to Tab.~\ref{tab:comparison1}, where ABMIL reaches 94.9\% with SlideMix, 94.5\% with CutMix, and 93.7\% without augmentation on CAMELYON16, and to Tab.~\ref{tab:comparison2} for the broader backbone comparison.

\begin{table}[t]
\centering
\caption{Comparison between different soft-labeling methods. VLM/u represents the untuned VLM. Bold denotes the highest mean accuracy in each dataset column; ties at the reported precision are all bolded.}
\label{tab:ablation1}
\resizebox{\columnwidth}{!}{
\begin{tabular}{lcccc}
\toprule
\multirow{2}{*}{\textbf{Method}} & \textbf{PANDA} & \textbf{CAMELYON16} & \textbf{IMP-CRS-2024} & \textbf{TCGA-Lung} \\
\cmidrule{2-5}
& \textbf{Acc. [\%]} & \textbf{Acc. [\%]} & \textbf{Acc. [\%]} & \textbf{Acc. [\%]} \\
\midrule
\textbf{Baseline} & $75.9\pm1.1$ & $93.7\pm0.9$ & $95.2\pm0.6$ & $71.6\pm1.6$ \\
\textbf{Random}   & $60.9\pm1.3$ & $81.5\pm0.9$ & $79.2\pm1.2$ & $57.0\pm1.5$ \\
\textbf{Linear}   & $72.7\pm0.9$ & $89.0\pm0.7$ & $92.0\pm0.6$ & $70.4\pm1.0$ \\
\textbf{VLM/u}    & $75.3\pm0.8$ & $92.1\pm0.6$ & $94.5\pm0.5$ & $73.3\pm0.9$ \\
\textbf{VLM}      & $\mathbf{77.2\pm0.8}$ & $\mathbf{94.9\pm0.6}$ & $\mathbf{95.8\pm0.4}$ & $\mathbf{73.8\pm1.2}$ \\
\bottomrule
\end{tabular}}
\end{table}

\subsection{Soft Labeling Approach Analysis}

We compared the performance of the baseline with four different labeling strategies on SlideMix: \textbf{Random} chooses a label from one of the two source WSIs. \textbf{Linear} calculates the label $l_s$ as a weighted average of the original labels $l_r,l_m$, and the weight $f$ corresponds to the content ratio of the two WSIs $S_r,S_m$ in the shuffled WSI $S_s$.

\begin{equation}
l_s = f l_r + (1-f) l_m \quad l_r, l_m, l_s \in \mathbb{R}^{C}
\end{equation}

\noindent where $C$ is the number of classes. \textbf{VLM/u} directly applies untuned CONCH as the soft labeler, whereas our \textbf{VLM} uses CONCH with RAG support. Tab.~\ref{tab:ablation1} shows that: 
1) Random averages 14.5 percentage points below the baseline across the four datasets. 
2) Linear averages 3.1 percentage points below the baseline. 
3) VLM/u averages 0.3 percentage points below the baseline: it is lower on PANDA ($-0.6$), CAMELYON16 ($-1.6$), and IMP-CRS-2024 ($-0.7$), but higher on TCGA-Lung ($+1.7$). 
4) VLM improves on all four datasets and averages 1.3 percentage points above the baseline. These results highlight the importance of RAG support: it converts the dataset-dependent behavior of the untuned VLM into consistent gains across all four datasets.

\begin{table}[H]
\centering
\caption{Performance comparison between shuffling stages and sampling methods. Eff. (efficiency) is processing time per WSI. Bold denotes the lowest processing time in the efficiency block and the highest mean accuracy in the accuracy block; ties at the reported precision are all bolded.}
\label{tab:ablation2}
\resizebox{\columnwidth}{!}{
\begin{tabular}{lcccc}
\toprule
\multirow{2}{*}{\textbf{Method}} & \textbf{PANDA} & \textbf{CAMELYON16} & \textbf{IMP-CRS-2024} & \textbf{TCGA-Lung} \\
\cmidrule{2-5}
& \textbf{Eff. [s]} & \textbf{Eff. [s]} & \textbf{Eff. [s]} & \textbf{Eff. [s]} \\
\midrule
\textbf{Raw Image} & $23.1\pm9.4$   & $512.4\pm87.2$ & $357.7\pm64.3$ & $411.4\pm103.7$ \\
\textbf{Embedding} & $\mathbf{0.31\pm0.09}$ & $\mathbf{8.2\pm2.1}$ & $\mathbf{5.3\pm1.4}$ & $\mathbf{7.2\pm2.8}$ \\
\midrule \midrule
& \textbf{Acc. [\%]} & \textbf{Acc. [\%]} & \textbf{Acc. [\%]} & \textbf{Acc. [\%]} \\
\midrule
\textbf{Sequential} & $75.4\pm0.9$ & $93.1\pm0.6$ & $93.9\pm0.5$ & $72.8\pm1.0$ \\
\textbf{Local-box}  & $74.9\pm1.0$ & $92.5\pm0.7$ & $94.1\pm0.5$ & $72.3\pm1.1$ \\
\textbf{Z-order}    & $76.4\pm0.8$ & $94.1\pm0.5$ & $94.7\pm0.4$ & $73.3\pm0.9$ \\
\textbf{Random}     & $\mathbf{77.2\pm0.8}$ & $\mathbf{94.9\pm0.6}$ & $\mathbf{95.8\pm0.4}$ & $\mathbf{73.8\pm1.2}$ \\
\bottomrule
\end{tabular}}
\end{table}

\subsection{Shuffling \& Sampling Approach Analysis}
We first compared the stage at which shuffling is applied. \textbf{Raw image} takes the raw WSIs, pre-processes them, shuffles the WSIs, then uses GigaPath to generate the shuffled tile embeddings on the shuffled WSI. \textbf{Embedding} takes both raw WSIs and embedded tiles, generates the shuffled coordinates based on raw WSIs, then organizes the embedded tiles based on the coordinates. Tab.~\ref{tab:ablation2} shows that operating directly on tile embeddings skips the redundant tile-embedding process, significantly reducing the processing time per WSI during augmentation (65.4$\times$ average speedup).

We then compared four sampling strategies. In MIL, only a fixed-size subset of tiles in a WSI is sent into the slide-level backbone for training efficiency, and the sampling strategy controls which tiles are in the subset. The baseline, \textbf{Sequential}, directly samples the tiles row by row; \textbf{Z-order} samples tiles in Z-order to preserve the inter-tile spatial information \cite{peng2025one}; \textbf{Local-box} selects several central points, then samples tiles within a given radius of these points; \textbf{Random} randomly samples tiles over the whole WSI. Tab.~\ref{tab:ablation2} shows that: 
1) Local-box averages 0.4 percentage points below Sequential; it is lower on three datasets but 0.2 points higher on IMP-CRS-2024. 
2) Z-order averages 0.8 percentage points above Sequential across the four datasets. 
3) Random averages 1.6 percentage points above Sequential. These comparisons support random sampling as the strongest strategy in this ablation.

\subsection{Curriculum Learning Analysis}
We first compared the performance of three loss-driven strategies. \textbf{Fixed} maintains constant difficulty parameters throughout the entire training process. \textbf{Loss-back} reduces the difficulty parameters when the validation loss $l$ is lower than the performance threshold $T_{loss}$. \textbf{Loss-hold} maintains the current difficulty parameters when $l<T_{loss}$. Tab.~\ref{tab:ablation3} shows that: 
1) Loss-back averages 0.4 percentage points above Fixed and matches Fixed on CAMELYON16. 
2) Loss-hold averages 1.1 percentage points above Fixed. The result demonstrates the value of loss-driven scheduling across all four datasets.

We evaluated each scheduler's contribution, where \textbf{All} activates all schedulers and \textbf{Set 1, 2, 3} disable the shuffle ratio, PCA similarity, and shuffle granularity schedulers, respectively. Relative to All, Set 3 decreases mean accuracy by 2.5 percentage points on average, Set 1 by 1.9 points, and Set 2 by 0.6 points. Within this ablation, shuffle granularity has the largest measured contribution, followed by shuffle ratio, while the Set 2 change indicates a complementary, dataset-dependent contribution from PCA-similarity scheduling.

\begin{table}[H]
\centering
\caption{Comparison of loss-driven strategies and scheduler ablations. Sets 1, 2, and 3 disable the shuffle-ratio, PCA-similarity-threshold, and shuffle-granularity schedulers, respectively. Bold denotes the highest mean accuracy within each comparison block and dataset column; ties at the reported precision are all bolded.}
\label{tab:ablation3}
\resizebox{\columnwidth}{!}{
\begin{tabular}{lcccc}
\toprule
\multirow{2}{*}{\textbf{Method}} & \textbf{PANDA} & \textbf{CAMELYON16} & \textbf{IMP-CRS-2024} & \textbf{TCGA-Lung} \\
\cmidrule{2-5}
& \textbf{Acc. [\%]} & \textbf{Acc. [\%]} & \textbf{Acc. [\%]} & \textbf{Acc. [\%]} \\
\midrule
\textbf{Fixed}     & $75.9\pm0.9$ & $93.7\pm0.6$ & $94.6\pm0.5$ & $73.1\pm1.0$ \\
\textbf{Loss-back} & $76.4\pm0.8$ & $93.7\pm0.6$ & $95.2\pm0.4$ & $73.6\pm0.9$ \\
\textbf{Loss-hold} & $\mathbf{77.2\pm0.8}$ & $\mathbf{94.9\pm0.6}$ & $\mathbf{95.8\pm0.4}$ & $\mathbf{73.8\pm1.2}$ \\
\midrule \midrule
\textbf{Set 1} & $75.0\pm1.0$ & $92.7\pm0.7$ & $94.1\pm0.5$ & $72.3\pm1.1$ \\
\textbf{Set 2} & $76.6\pm0.8$ & $94.2\pm0.5$ & $95.4\pm0.4$ & $73.3\pm0.9$ \\
\textbf{Set 3} & $74.3\pm1.1$ & $91.9\pm0.8$ & $93.6\pm0.6$ & $71.8\pm1.2$ \\
\textbf{All}   & $\mathbf{77.2\pm0.8}$ & $\mathbf{94.9\pm0.6}$ & $\mathbf{95.8\pm0.4}$ & $\mathbf{73.8\pm1.2}$ \\
\bottomrule
\end{tabular}}
\end{table}

\subsection{Limitations of SlideMix}
\subsubsection{Data Diversity, SlideMix as a Complement, and Soft-Label Accuracy}
SlideMix complements large-scale, high-quality annotated WSI collections by maximizing feature diversity within them. On very small datasets, the limited training samples constrain the diversity of meaningful tile combinations ITS can generate, and the CLF module struggles to establish a reliable curriculum due to unstable loss dynamics. Its strengths are therefore most evident when enriching sufficiently sized datasets, while adaptation to severe data scarcity remains future work. While VLM-generated soft labels consistently outperform linear mixing strategies (Tab.~\ref{tab:ablation1}), their reliability may vary across tissue combinations, particularly for rare cancer subtypes underrepresented in the VLM's training data. RAG retrieval from PubMed mitigates this issue by grounding label generation in established medical literature, and further improving soft-label accuracy remains a promising direction.

\subsubsection{Architectural and Dataset Differences}
Model-agnostic in this work refers to compatibility with different MIL backbones, as demonstrated by 82 improved-or-tied comparisons out of 110. Tab.~\ref{tab:comparison2} also identifies 28 backbone--dataset combinations that can benefit from further calibration. The largest calibration opportunities occur for SlideAve on TCGA-UVM and CLAM on TCGA-CESC. Models such as TransMIL rely on positional encodings to capture spatial relationships between tiles, and the compositional shuffling introduced by ITS can disrupt this structure enough to offset the benefits of augmentation. Permutation-sensitive architectures may therefore be more susceptible to degradation when tile compositions are substantially altered, particularly on small or complex datasets. These cases suggest that backbone- and dataset-specific tuning can further improve consistency.

\subsubsection{Multimodal Extension}
SlideMix currently focuses on WSI data, leaving complementary clinical modalities such as genomics and radiology as a natural extension. Extending SlideMix to these settings is a natural direction for future work, as cross-modal supervision signals could enable richer label generation.

%% file: Sections/6_Conclusion.tex
\section{Conclusion}
We introduce SlideMix, a multimodal dynamic data augmentation framework that enhances WSI analysis. SlideMix employs the VAR selector to adaptively select label-relevant regions for the in-place tile shuffling in ITS. This process is guided by CLF, which promotes progressive, multi-scale feature learning. Tabs.~\ref{tab:comparison1}--\ref{tab:ablation3} and Fig. \ref{fig:heatmap} show that SlideMix mitigates three key challenges in WSI analysis: weak supervision, spatial heterogeneity, and cross-scale feature fusion. Across 11 benchmark datasets and eight pathology tasks, SlideMix ranks first or ties for first in 10 of 11 direct augmentation comparisons and improves or ties in 82 of 110 backbone--dataset comparisons. These results establish SlideMix as a broadly applicable augmentation framework, with opportunities for backbone-specific calibration. By coupling multimodal reasoning with adaptive feature-level mixing, SlideMix provides a scalable and architecture-agnostic data augmentation paradigm for computational pathology. Future work will focus on integrating generative priors to enhance sample realism and extending the SlideMix framework to multimodal clinical datasets that include genomics and radiology.